\documentclass[10pt,conference]{IEEEtran}
\IEEEoverridecommandlockouts
\usepackage{hyperref}

\usepackage{textcomp}
\usepackage{xcolor}
\usepackage{array}
\usepackage{rotating}
\usepackage{multirow}
\usepackage{float} 
\usepackage{booktabs}
\usepackage{url}
\usepackage{verbatim}
\usepackage{graphicx}
\usepackage[font={footnotesize}]{subcaption}
\usepackage{float}
\usepackage{amsmath}
\usepackage{amsthm}
\usepackage{amssymb}
\usepackage{setspace}
\usepackage{tabularx}
\usepackage{tikz}
\usepackage{xcolor}
\usepackage{cite}

\usepackage{enumitem}

\newcommand{\argmin}[1]{\underset{#1}{\operatorname{arg}\,\operatorname{min}}\;}

\begin{document}

\title{Overcoming Noisy and Irrelevant Data in\\ Federated Learning \vspace{-0.05in}}

\author{
\IEEEauthorblockN{Tiffany Tuor\IEEEauthorrefmark{1}, Shiqiang Wang\IEEEauthorrefmark{2}, Bong Jun Ko\IEEEauthorrefmark{3}, Changchang Liu\IEEEauthorrefmark{2}, Kin K. Leung\IEEEauthorrefmark{1}}
\IEEEauthorblockA{\IEEEauthorrefmark{1}Department of Electrical and Electronic Engineering, Imperial College London, UK}
\IEEEauthorblockA{\IEEEauthorrefmark{2}IBM T. J. Watson Research Center, Yorktown Heights, NY, USA}
\IEEEauthorblockA{\IEEEauthorrefmark{3}Stanford Institute for Human-Centered Artificial Intelligence (HAI), Stanford, CA, USA}
\IEEEauthorblockA{Email: \{tiffany.tuor14, kin.leung\}@imperial.ac.uk, \{wangshiq@us., Changchang.Liu33@\}ibm.com, bongjun@gmail.com}
\thanks{
This paper has been accepted in the 25th International Conference on Pattern Recognition (ICPR).

This research was sponsored by the U.S. Army Research Laboratory and the U.K. Ministry of Defence under Agreement Number W911NF-16-3-0001. The views and conclusions contained in this document are those of the authors and should not be interpreted as representing the official policies, either expressed or implied, of the U.S. Army Research Laboratory, the U.S. Government, the U.K. Ministry of Defence or the U.K. Government. The U.S. and U.K. Governments are authorized to reproduce and distribute reprints for Government purposes notwithstanding any copyright notation hereon.}
\vspace{-0.25in}
}

\maketitle

\begin{abstract}
Many image and vision applications require a large amount of data for model training. Collecting all such data at a central location can be challenging due to data privacy and communication bandwidth restrictions. Federated learning is an effective way of training a machine learning model in a distributed manner from local data collected by client devices, which does not require exchanging the raw data among clients. A challenge is that among the large variety of data collected at each client, it is likely that only a subset is relevant for a learning task while the rest of data has a negative impact on model training. Therefore, before starting the learning process, it is important to select the subset of data that is relevant to the given federated learning task. In this paper, we propose a method for distributedly selecting relevant data, where we use a benchmark model trained on a small benchmark dataset that is task-specific, to evaluate the relevance of individual data samples at each client and select the data with sufficiently high relevance. Then, each client only uses the selected subset of its data in the federated learning process. The effectiveness of our proposed approach is evaluated on multiple real-world image datasets in a simulated system with a large number of clients, showing up to $25\%$ improvement in model accuracy compared to training with all data.
\end{abstract}


\section{Introduction}
\label{sec:intro}

Modern applications of image recognition and computer vision are usually based on machine learning models such as deep neural networks. Training such models requires a large amount of data that are specific to the learning task. Due to privacy regulations and communication bandwidth limitation, it can be difficult to collect all the data at a central location.
To benefit from the local data collected by and stored at client devices (e.g., mobile phones, surveillance cameras), \emph{federated learning} has been proposed as a way of distributed model training without exchanging raw data, which preserves data privacy and conserves communication bandwidth~\cite{mcmahan2016communication,park2019wireless,yang2019federated,li2019federated,kairouz2019advances}. In federated learning, multiple clients participate in model training in a collaborative and coordinated manner. The learning proceeds by iteratively exchanging model parameters between the clients and a server, while each client's local data are \emph{not} shared with others.

Existing work on federated learning mostly focused on model training  using pre-defined datasets at client devices. While such scenarios are meaningful, it can be far from practical situations where each client can have a large variety of data which may or may not be relevant to the given federated learning task, and the relevance is unknown by the system. For example, if the task is to classify handwritten digits, printed digits can be considered as irrelevant data although they may be tagged with the same set of labels (thus difficult for the system to distinguish). Including irrelevant data in the federated learning process can reduce model accuracy and slow down training.
In this paper, we address the following problem: how does a client select relevant data for a given federated learning task?

We consider a federated learning system with a \emph{model requester} (MR), multiple clients, and a server. A federated learning task is defined by the MR which provides the model logic (e.g., a specific deep neural network architecture). The MR can be any user device with computational capability, which may or may not be a client or the server at the same time.

We propose a novel method for identifying relevant data at each client for a given federated learning task. In our proposed approach, the MR has a small set of \emph{benchmark data} that is used as an example to capture the desired input-output relation of the trained model, but is insufficient to train the model itself. The benchmark data does \emph{not} need to be shared with the server or other clients directly, only a \emph{benchmark model} trained on the benchmark data needs to be shared, thus preserving privacy. Using the benchmark model provided by the MR, each client identifies a subset of its data that will be involved in this federated learning task. Then, federated learning proceeds, where each client's local computation is only performed on its selected data. In this way, our approach works in a distributed manner without requiring clients or MR to share raw data.

The performance of our proposed approach is evaluated with extensive experiments. On a variety of real-world image datasets with different types of noise (see the definition of ``noise'' in Section~\ref{sec:existing-work}), we show that our data selection approach outperforms other baseline approaches even when the MR only has a very small amount of benchmark data. Furthermore, we show that data selection also speeds up the overall federated learning process.

\section{Related Work}
\label{sec:existing-work}

Federated learning was first introduced by \cite{mcmahan2016communication}, which allows multiple client devices to learn a global model while keeping their raw data stored locally. In federated learning settings, some work has studied the case with malicious clients~\cite{blanchard2017machine,fung2018mitigating} and servers~\cite{wang2019beyond}, where both defense and attacking mechanisms are considered.
The above body of work focuses on a unified behavior of a client or the server in its entirety, i.e., the entire client/server is either malicious or not.  
It does not allow the more realistic setting where relevant and irrelevant data (with respect to a particular learning task) coexist at each client.

Relevant and irrelevant data in the training dataset has been considered for centralized machine learning in recent years particularly in the domain of image recognition, where irrelevant data can be regarded as \emph{noise} for a given machine learning task. Note that, however, noise for one task may be useful data for another task.
Different categories of noise have been identified by \cite{wang2018iterative} for supervised learning tasks with a fixed number of (known) classes:
\begin{enumerate}
    \item Closed-set vs. open-set noise 
    \begin{itemize}
        \item \emph{Closed-set noise}: Data sample of a known class is labeled as another known class.
        \item \emph{Open-set noise}: Data sample of an \emph{unknown} class is labeled as a known class\footnote{For example, to classify images of cats of dogs, an image of a dog labeled as a cat is a closed-set noise, an image of an elephant labeled as a dog is an open-set noise, because elephant does not belong to the set of known classes (cats and dogs).}.
    \end{itemize}
    \item Strong vs. weak noise 
    \begin{itemize}
        \item \emph{Strong noise}: The number of noisy data samples can exceed the number of clean data samples.
        \item \emph{Weak noise}: The number of noisy data samples  is less than the number of clean data samples.
    \end{itemize}
\end{enumerate}

Among these different types of noise, approaches for model training in the \emph{weak closed-set} noise setting has been most extensively studied~\cite{patrini2017making,han2018co,ghosh2017robust}, and the \emph{strong closed-set} noise setting has been studied by \cite{veit2017learning}, \cite{vahdat2017toward}, and \cite{hendrycks2018using}. Very recently, the \emph{weak open-set} noise setting is studied by \cite{wang2018iterative}, \cite{yu2019does}, and \cite{lee2019robust}.
We note that all the approaches that address the strong noise scenario require a small set of clean benchmark data, as we do in our paper. In addition, existing approaches for the open-set setting involve very complex models that need to be trained on the entire dataset even though they only apply to the weak noise scenario, which makes them infeasible for distributed datasets in federated learning.

The problem of filtering out noisy training data is also broadly related to the area of anomaly/outlier detection~\cite{Bergman2020Classification-Based}. However, anomaly detection techniques focus on detecting anomalous data samples based on the underlying distribution of the data itself, instead of how the data impacts model training. For example, the impact of two training data samples may be similar for one learning task, but different for another learning task. Such task-specific characteristics are not considered in anomaly/outlier detection techniques.

In summary, there is a significant gap between the above existing work and the problem we solve in this paper. To the best of our knowledge, none of the following aspects has been studied in the literature: 1) strong open-set noise in training data and its impact on the learning performance, 2)~data cleansing/filtering in federated learning with decentralized datasets located at clients.
In federated learning, strong open-set noise can frequently exist in scenarios where only a small portion of the data at each client is relevant to the given task. In this paper, we propose a novel method of filtering the data in a \emph{distributed manner} for federated learning with \emph{strong open-set noise}, which also works in weak noise and closed-set noise scenarios at the same time.

The remainder of this paper is organized as follows. Section~\ref{section:federated} presents the system model and basic algorithm of federated learning. Section~\ref{sec:data-filter} describes our proposed data selection method. The experimentation results are presented in Section~\ref{sec:experimentation} and Section~\ref{sec:conclusion} draws conclusion.

\section{System Model and Definitions}
\label{section:federated}

We consider a federated learning system with multiple clients and a server. Each client has its own local dataset with diverse types of data. As described in Section~\ref{sec:intro}, a federated learning task is initiated by an MR. The MR specifies what model needs to be trained, and also has a small \emph{benchmark dataset} that can be generated by the MR based on its own knowledge of the learning task or via other means, but are deemed to have correct labels for all data in it. In the case where the MR is also a client, this client's local dataset can be considered as the benchmark data. However, the amount of benchmark data is usually very small compared to the collection of all clients' local data and is therefore insufficient for training a highly accurate model, which is why federated learning involving multiple clients is needed.

Each client may have different amounts of relevant and irrelevant data with respect to a federated learning task. After the system receives the model training request, a distributed data selection procedure is performed, which is described in details in Section~\ref{sec:data-filter}. The selected relevant data at each client are then used in the federated learning process of this task.

In this paper, we assume that the clients are trusted and cooperative. They voluntarily participate in refining the dataset for each task. The noise in the data (with respect to the task) can be due to the following reasons: 1) irrelevance of the data to the task, 2) data mislabeling by the client or by a third-party that provides data to the client, 3) lack of proper representation of data labels in the system (e.g., a label ``0'' can represent different things depending on the context).

We consider a set of data $\mathcal{D}$ (potentially including noisy data samples) distributed over $N$ clients such that  $\mathcal{D}=\bigcup_{n=1}^{N} {\mathcal{D}_n} $, where ${\mathcal{D}_n}$ is the dataset located at client $n$. We focus on supervised learning with labeled training data in this paper.
The \emph{loss function} of a labeled data sample $(x_i, y_i) \in \mathcal{D}$ is defined as $l(f(x_i,\theta),y_i)$, where $x_i$ is the input to the model, $y_i$ is the desired output (label) of the model,  $f(x_i,\theta)=\hat y_i$ is the predicted output of the model, and $\theta$ is the model parameter vector. The function $f(x_i,\theta)$ captures the model logic and can be different for different models. The loss function $l(f(x_i,\theta),y_i)$ is an error function in the form of mean squared error, cross entropy, etc. 

For a federated learning task, let $ \mathcal{F}\subseteq \mathcal{D}$ denote the set of samples that is selected as relevant to the given task. This set is found by finding the subset  $\mathcal{F}_n \subseteq \mathcal{D}_n$ in each client $n$ ($ \mathcal{F}=\bigcup_{n=1}^{N}{\mathcal{F}_n}$) and the details of this selection process will be given in Section~\ref{sec:data-filter}.
The overall loss of relevant data at client $n$ is defined as 
\begin{equation}
L_n(\theta)=
\frac{1}{|\mathcal{F}_n|}\sum_{(x_i,y_i)\in \mathcal{F}_n} l(f(x_i,\theta),y_i)
\end{equation}
where $|\cdot|$ denotes the cardinality of the set, based on which the global loss across all clients is defined as 
\begin{equation} L(\theta)=\frac{\sum_{n=1}^{N} |\mathcal{F}_n|L_n(\theta)}{\sum_{n=1}^{N} |\mathcal{F}_n|}.
\end{equation}
The goal of federated learning on the selected data subset $\mathcal{F}$ is to find the model parameter vector $\hat{\theta}$ that minimizes $L(\theta)$:
\begin{equation}
\hat\theta=\argmin{\theta} L(\theta).
\label{eq:MLObjective}
\end{equation}

The minimization problem in (\ref{eq:MLObjective}) is solved in a distributed manner using standard federated learning procedure~\cite{mcmahan2016communication,wang2019adaptive} that includes the following steps:

\begin{enumerate}
    \item Each client $n$ (simultaneously)  performs $\tau \geq 1$ steps of stochastic gradient descent (SGD) on its local model parameter $\theta_i$, according to
    \begin{equation}
    \label{eq:local}
    \theta_n(t)= \theta_n(t-1)-\eta \nabla L_n(\theta_n(t-1))
    \end{equation}
    where we consider the number of local iteration steps (between communications with the server) $\tau$ as a fixed parameter in this paper, and $\nabla L_n(\theta_n(t-1))$ is the stochastic gradient of the loss value computed on a \emph{mini-batch} of data randomly sampled from $\mathcal{F}_n$.
    
    \item  Each client $n$ sends its new parameter $\theta_n(t)$ to the server.
    
    \item The server aggregates the parameters received from each client, according to
    \begin{equation}
    \label{eq:global-agg}
    \theta(t)= \frac{\sum_{n=1}^{N} |\mathcal{F}_n|\theta_n (t)}{\sum_{n=1}^{N} |\mathcal{F}_n|} .
    \end{equation}
    
    \item The server sends the global model parameter $\theta(t)$ computed in (\ref{eq:global-agg}) to each client. After receiving the global model parameter, each client $n$ updates its local model parameter $\theta_n(t) \leftarrow \theta(t)$.
     
\end{enumerate}
The above steps repeat until training convergence.
The size of the mini-batch at client $n$ to compute $\nabla L_n(\theta_n(t-1))$ should be proportional to $|\mathcal{F}_n|$, to ensure that the stochastic gradient is an unbiased estimator of the actual gradient~\cite{tuor2018distributed} and that the number of iterations for a full pass through the data (i.e., an epoch) in $\mathcal{F}_n$ is the same for each client $n$. In this paper, we use a fixed percentage of $|\mathcal{F}_n|$ as the mini-batch size of each client $n$, thus different clients may have different mini-batch sizes depending on  $|\mathcal{F}_n|$.

\section{Data Selection}
\label{sec:data-filter}

Data selection is performed at the beginning of each federated learning task, before the distributed model training process (i.e., Steps 1--4 in Section~\ref{section:federated}) starts. The goal is to filter out noisy data samples to the specific federated learning task. The filtering process is decentralized, where each client performs the filtering operation on its own dataset and only exchanges a small amount of meta-information with the server.

The main idea is that the MR first trains a \emph{benchmark model} (of the same type as the final model requested by the MR) using \emph{only a subset} of the benchmark data owned by the MR\footnote{Note that training the benchmark model is usually less computationally intensive than training the final model using federated learning, because the benchmark dataset size is usually small. Therefore, we assume that the benchmark model can be trained by the MR on its own.}.
Although the benchmark model is trained with a small amount of data that is insufficient to provide a good classification accuracy on its own, it still captures useful characteristics of the training data, which is helpful for determining the relevance of local data at clients. In particular, for a data sample that is irrelevant to the learning task, the benchmark model would likely predict a label distribution (i.e., the output of the model) that has a higher entropy. In other words, the predicted label is more uncertain for irrelevant data samples. Such uncertainty (or incorrectness of the predicted label) is reflected by a higher loss value for the particular sample.

Then, to determine relevance, each client $n$ evaluates the loss $l(f(x_i,\theta),y_i)$ for each data sample $(x_i, y_i) \in \mathcal{D}_n$ using the benchmark model. By comparing the distribution of loss values of clients' data samples and a held-out set of benchmark data (that was not used for training the benchmark model), we determine a threshold; data samples with loss higher than the threshold are seen as noise and excluded from training.


\begin{figure}
  \centering
     \includegraphics[width=1\columnwidth]{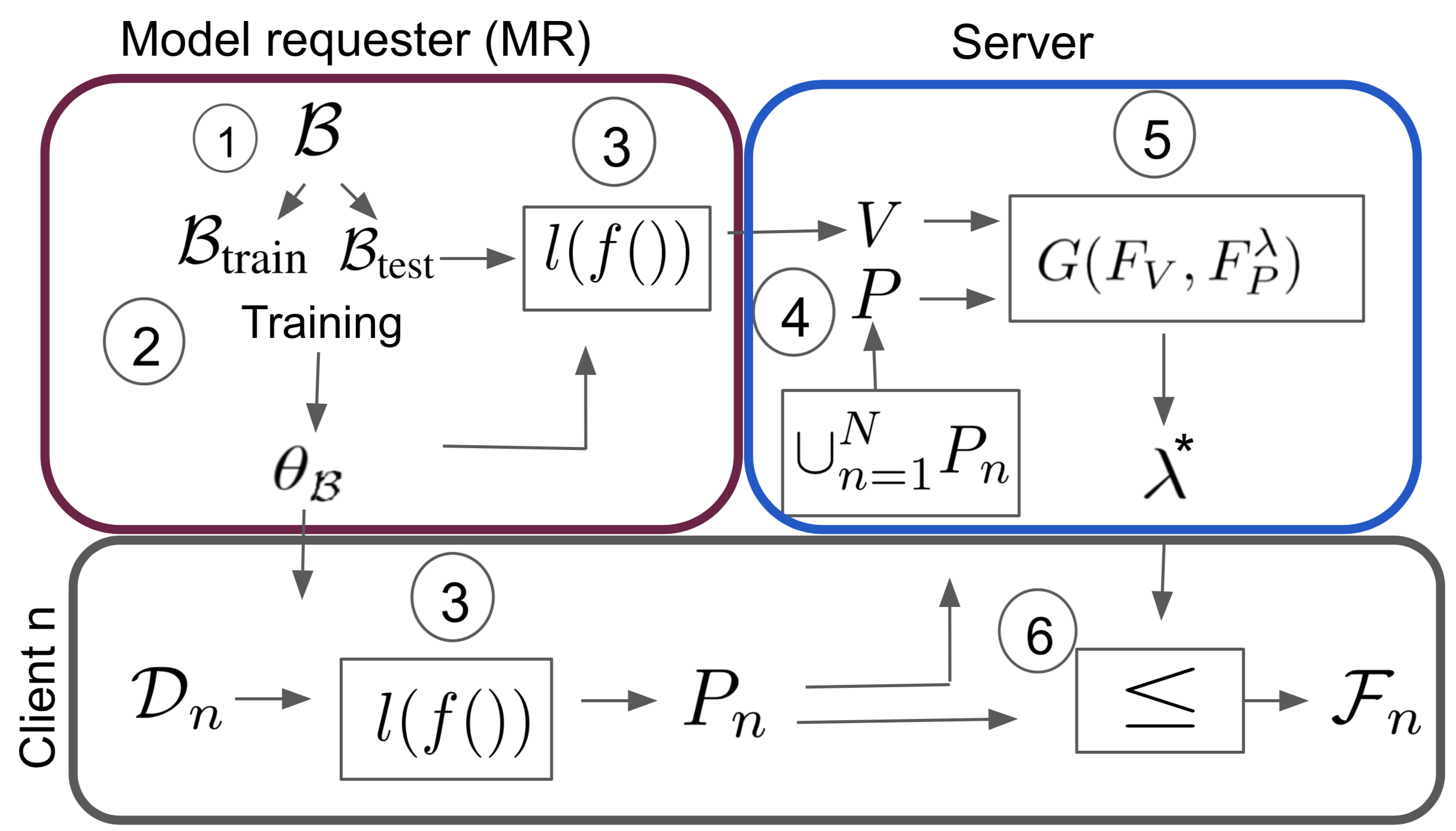}
    \caption{Data selection procedure.}
    \label{fig:preprocessing}
\end{figure}

Figure~\ref{fig:preprocessing}  shows the overall data selection process involving the MR, server, and clients, including the following steps:

\begin{enumerate}[label=\alph*)]
    \item The MR builds (trains) the benchmark model using a small benchmark dataset without noise (Steps 1 and 2). It then calculates a distribution of loss values given by the benchmark model on the held-out benchmark data.
    \item Each client evaluates its own dataset against the benchmark model, and creates a set of loss values computed on all its data samples (Step 3). 
    \item The server merges the sets of loss values from all clients, and compares the distribution of clients' loss values against the distribution of loss values of the benchmark dataset (computed by the MR), to calculate the filtering threshold (Steps 4 and 5).
    \item Each client filters out noisy data samples in its dataset using the filtering threshold (Step 6).
\end{enumerate}

Note that this process is performed \emph{without transmitting raw data} at the MR/clients to the server; only a benchmark model and the losses are exchanged, hence suitable for federated learning settings. In the following, we begin the detailed description of this filtering process by formally establishing its objective.

\subsection{Objective}

Assume an MR wants to train a classifier able to recognize the classes (labels) defined in a label set $C$ and has a set of benchmark data ${\mathcal{B}}$ which only contains a small number of examples correctly labeled from each of the categories defined in $C$; typically $|\mathcal{B}| \ll |\mathcal{D}|$. For example, an MR whose focus is on recognizing images of \textit{cats} and \textit{dogs} will provide a few correctly labeled examples representing \textit{cats} and \textit{dogs}, and $C=\{\text{dog}, \text{cat}\}$ is implicitly defined. 

Our overall goal of data selection is to find the subset $ \mathcal{F}\subseteq \mathcal{D}$, such that the \emph{testing} loss function evaluated on classes defined by $C$ is minimized: 
\begin{equation}
\begin{array}{rrclcl}
\displaystyle \min_{\mathcal{F}} & \multicolumn{3}{l}{\sum_{(x_i,y_i)\in \mathcal{T}} l(f(x_i,\hat\theta);y_i)}
\end{array}
\label{eq:minization_d}
\end{equation}
where the test dataset $\mathcal{T}=\{(x_i,y_i) : y_i \in C \}$ is a held-out dataset, separated from the training dataset $\mathcal{D}$ and the benchmark dataset $\mathcal{B}$ (i.e., $\mathcal{T} \cap \mathcal{D} = \emptyset$ and $\mathcal{T} \cap \mathcal{B} = \emptyset$), and $\hat{\theta}$ is the parameter vector of the model obtained from (\ref{eq:MLObjective}).

\begin{figure}
  \centering
    \includegraphics[width=1\columnwidth]{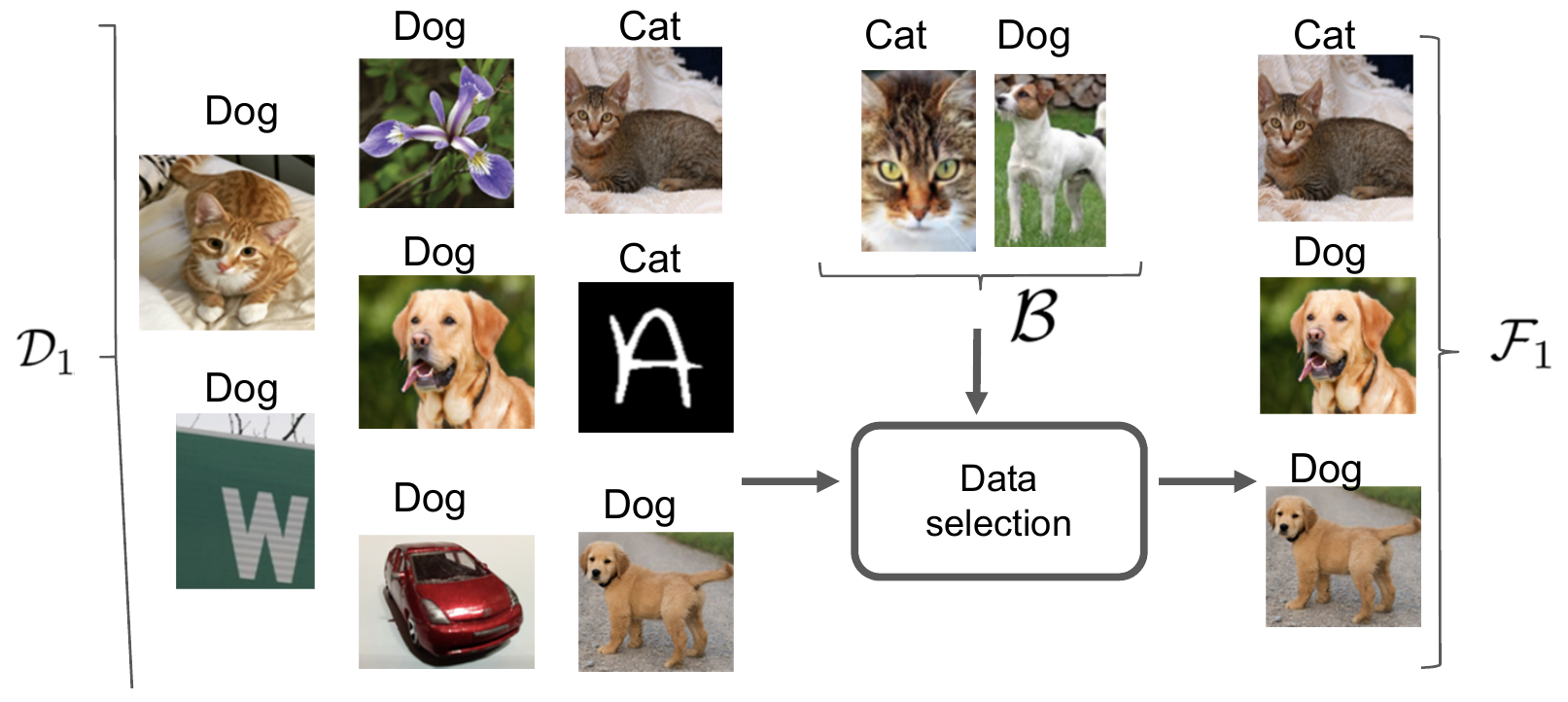}
    \caption{Example of noisy data at clients ($\mathcal{D}_1$), benchmark dataset ($\mathcal{B}$), and the filtered dataset ($\mathcal{F}_1$).}
    \label{fig:image-example}
\end{figure}

Figure~\ref{fig:image-example} shows an example of a noisy dataset ${\mathcal{D}_{1}}$, a benchmark dataset $\mathcal{B}$ owned by the MR, and the ideal subset $\mathcal{F}_1$, obtained using our proposed filtering method.

\subsection{Training Benchmark Model}

First, the benchmark dataset $\mathcal{B}$ is divided into a training set $\mathcal{B}_\textrm{train}$ and a test set $\mathcal{B}_\textrm{test}$ (Step 1 in Figure~\ref{fig:preprocessing}) such that $\mathcal{B}_\textrm{train} \cap\mathcal{B}_\textrm{test}  = \emptyset$. Then, the benchmark model is trained using $\mathcal{B}_\textrm{train}$, whose parameter $\theta_{\mathcal{B}}$ is obtained by minimizing the following expression using a model training process, e.g., SGD (Step~2):
\begin{equation}
\begin{array}{rrclcl}
\displaystyle \theta_{\mathcal{B} }=\argmin{\theta} &  \multicolumn{3}{l}{\frac{1}{|\mathcal{B}_\textrm{train}|} \sum_{(x_i,y_i)\in \mathcal{B}_\textrm{train} } l(f(x_i,\theta),y_i)}.
\\
\end{array}
\end{equation}

\subsection{Finding Loss Distribution Using Benchmark Model}

The model parameter $\theta_{\mathcal{B}}$ is used to find two sets of loss values (Figure~\ref{fig:preprocessing}, Step 3), evaluated on $\mathcal{\mathcal{B}}_\textrm{test}$ and $\mathcal{D}_n$ respectively:
\begin{align}
V &=\{l(f(x_i,\theta_{\mathcal{B} }),y_i) : \forall (x_i,y_i) \in \mathcal{B}_\textrm{test}\}
\label{eq:V} \\
P_n &=\{l(f(x_i,\theta_{\mathcal{B} }),y_i) : \forall (x_i,y_i) \in {\mathcal{D}_n}\}.
\label{eq:Pi}
\end{align}
Note that $V$ is obtained by the MR from $\mathcal{B}_\textrm{test}$, and each $P_n$ is obtained from $\mathcal{D}_n$ by each client $n$. Then, $V$ and all $P_n$'s are sent to the server, and $\{P_n\}$ is combined to obtain the loss value set of all clients' data (Step 4), i.e., $P =\bigcup_{n=1}^{N} P_n$.

The set $V$ provides a reference distribution of loss values, against which the relevance of the data samples at each client $n$ is assessed. Intuitively, we assume that the smaller the individual loss value $l(f(x_i,\theta_{\mathcal{B} }),y_i)$ is, the more $(x_i,y_i)$ will fit to the benchmark model defined by $\theta_{\mathcal{B}}$, hence being more likely relevant to the learning task under consideration. 

\subsection{Calculating Filtering Threshold in Loss Values}

\begin{figure}
    \centering
    \begin{subfigure}[b]{0.4\columnwidth}
        \centering
        \includegraphics[width=0.9\linewidth]{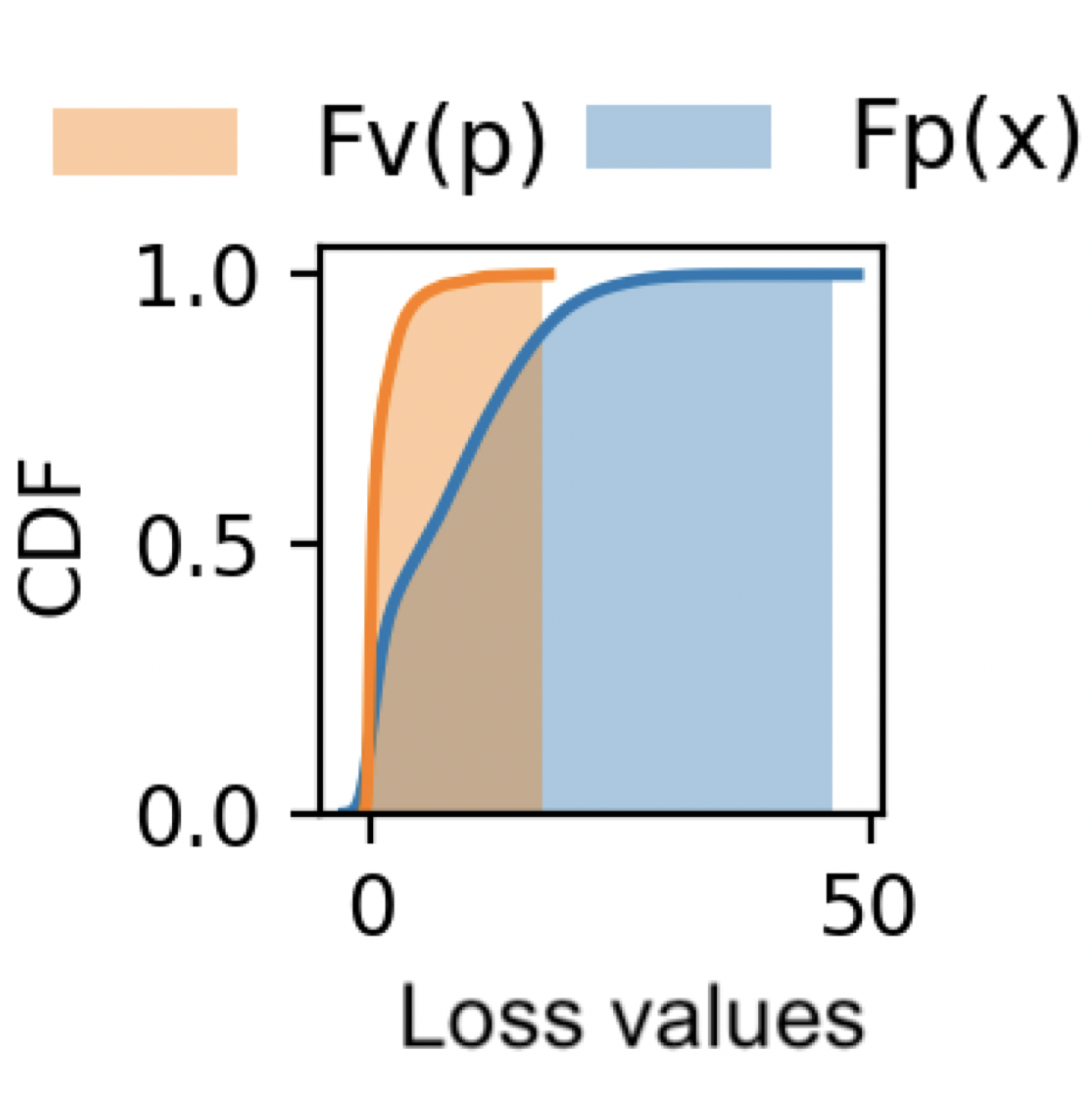}
        \caption{Original $F_V(x)$ and $F_{P}(x)$} 
        \label{fig:distance2}
    \end{subfigure}%
      ~~~~~~~
    \begin{subfigure}[b]{0.4\columnwidth}
        \centering
        \includegraphics[width=1 \linewidth]{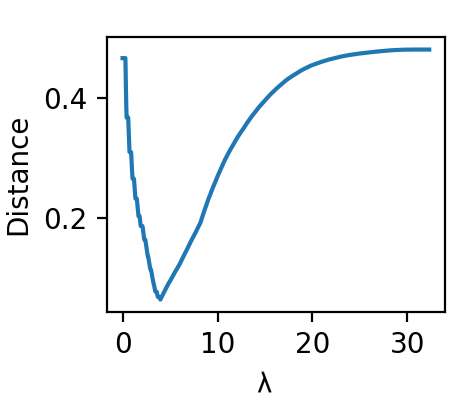}
        \caption{Distance for different  $\lambda$ }
        \label{fig:distance0}
    \end{subfigure}\vspace{0.1in}
    
    \begin{subfigure}[b]{0.3\columnwidth}
        \centering
        \includegraphics[width=1.3\linewidth]{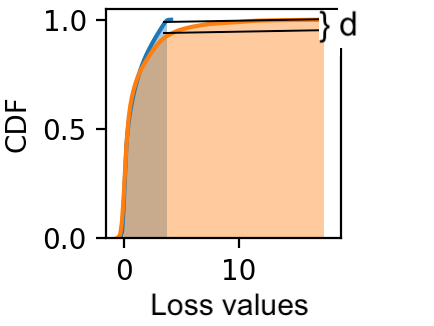}
        \caption{ $\lambda=4.8$} 
        \label{fig:distance3}
    \end{subfigure}
    ~
    \begin{subfigure}[b]{0.3\columnwidth}
        \centering
        \includegraphics[width=1.3\linewidth]{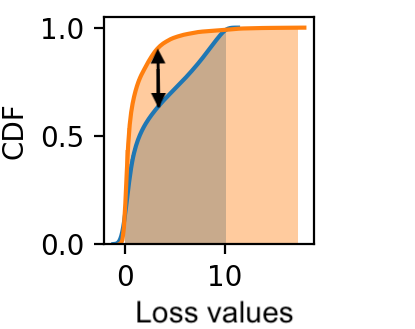}
        \caption{ $\lambda=10$ }
        \label{fig:distance4}
    \end{subfigure}%
        ~
    \begin{subfigure}[b]{0.3\columnwidth}
        \centering
        \includegraphics[width=1.3\linewidth]{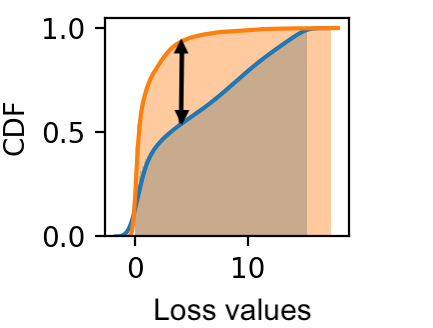}
        \caption{ $\lambda=15$ }
        \label{fig:distance5}
    \end{subfigure}%

\caption{KS distance and optimal $\lambda$ for $F_V(x)$ and $F^{\lambda}_{P}(x)$.}
\label{fig:distance}
\end{figure}

We use $V$ as a benchmark distribution of acceptable range of individual loss values for data samples in $\mathcal{D}$. In other words, we assume that data samples, whose loss values evaluated with the benchmark model are within an ``acceptable'' range in the distribution of $V$, have a high probability to be relevant to the learning of the target model. Inversely, if a sample has a loss value out of this acceptable range, there is a high probability that this sample will either corrupt the model training or just be irrelevant.

The goal is then to find a good threshold in the loss value to determine which data samples to include in the set of relevant (not noisy) data $\mathcal{F}_n$. Note that this can be seen as outlier detection, where the outliers are defined as the irrelevant data samples with respect to the target model, and their detection is performed in the 1-dimensional space of loss values mapped from the original data space via the benchmark model. Note also that we defined $V$ as the set of loss function values evaluated on $\mathcal{B}_\textrm{test}$ and not $\mathcal{B}_\textrm{train}$, to avoid loss values over-fitted to the subset of benchmark data used for training.

\begin{table*}[t]
\caption{Summary of datasets used in experiments \vspace{-0.1in}} {
\begin{center}
\small
\begin{tabular}{lccccl}
\hline
\textbf{} Dataset & \# Training & \# Testing   & Category & Format & Description \\ \hline
\textbf{MNIST}& 60,000 & 10,000 & 10 & $28\times28\times1 $  & hand-written digits  \\
\textbf{SVHN}& 73,257 &  26,032  & 10  & $32\times32\times3$ &  cropped digits from street view\\
\textbf{FEMNIST}&  71,090 &  8,085   & 62  &  $28\times28\times1 $ &  hand-written characters (federated learning setting)\\
\textbf{Chars74K}&   58,097 &  17,398 & 62 &  $28\times28\times1 $ & natural images, hand-written and font characters\\
\textbf{FASHION}& 60,000 & 10,000  & 10  & $28\times28\times1 $ &  fashion items\\
\textbf{CIFAR-10} & 50,000 & 10,000  & 10 &  $32\times32\times3 $ &  various object classes\\
\textbf{CIFAR-100}& 50,000 & 10,000  & 100 &  $32\times32\times3$ &  various object classes\\ \hline
\end{tabular}
\end{center}
}
\label{table:dataset}
\end{table*}

Our approach of detecting the outliers (i.e., noisy data) is to use $V$ as a mask to find an upper limit of acceptable loss values via a statistical test that compares the distribution of $V$ and $P$. More specifically, let us denote the empirical CDF (Cumulative Distribution Function) of $V$ and $P$ by $F_V$ and $F_P$, respectively; that is, $F_V (x) = \Pr\{X \le x: X \in V\}$ and $F_P (x) = \Pr\{X \le x: X \in P\}$. We further denote by $F^\lambda_P$ the conditional CDF of $P$ such that $F^\lambda_P (x) = \Pr\{X \le x | X \le \lambda : X \in P\}$. Note that, we call this conditional CDF the ``truncated'' CDF as the maximum range of values is truncated down to $\lambda$, i.e., $F^\lambda_P(x) = 1$ if $x \ge \lambda$, and $F^\lambda_P(x) = F_P(x) / F_P(\lambda)$ if $x < \lambda$.

Given $\lambda$, we define the distance $G$ between the two distributions $F_V$ and $F^\lambda_P$ by the Kolmogorov-Smirnov (KS) distance \cite{massey1951kolmogorov}, which is often used to quantify the distance between two CDFs. Specifically, $G (F_V, F^\lambda_P) = \sup_x |F_V(x) - F^\lambda_P (x)|$.

Then, we calculate our threshold in loss values, denoted by $\lambda^*$, that minimizes $G$, that is:
\begin{equation}
\lambda^* = \argmin{\lambda} G (F_V, F^\lambda_P).
\label{eq:tauOptimal}
\end{equation}

Figure~\ref{fig:distance} illustrates this process with an example.  Given the unconditional CDFs $F_V(x)$ and $F_{P}(x)$ in Figure~\ref{fig:distance2}, the distances between $F_V(x)$ and the truncated CDF $F^\lambda_{P}(x)$  for different values of $\lambda$ are shown in Figures~\ref{fig:distance3}, \ref{fig:distance4}, and \ref{fig:distance5}. 
In this example, $\lambda^* \approx 4.8$ because it minimizes the KS distance between $F_V$ and $F^\lambda_P$, as shown in Figure~\ref{fig:distance0}.

\subsection{Local Selection of Data by Clients}
After computing $\lambda^*$ according to (\ref{eq:tauOptimal}), the server sends $\lambda^*$ to all the clients. Then, each client $n$ makes the selection of relevant data locally (Step 6 in Figure~\ref{fig:preprocessing}):
\begin{equation}
\mathcal{F}_n=\{(x_i,y_i) \in \mathcal{D}_n :  l(f(x_i,\theta_{\mathcal{B} }),y_i) \leq \lambda^* \}.
\end{equation}

Once the selection is made locally for every client, the standard federated learning process starts, where each client~$n$ performs SGD on the selected data $\mathcal{F}_n$. As explained in Section~\ref{section:federated}, the mini-batch size is adapted according to the size of $\mathcal{F}_n$. In the extreme case where a client has no data of interest to a federated learning task, the mini-batch size for this client will be $0$ and this client is excluded from the federated learning task under consideration.

\section{Experimentation Results}
\label{sec:experimentation}

\subsection{Setup}
We conduct experiments in a federated learning system with a large number of simulated clients ($N \geq 100 $).

\subsubsection{Datasets}

We use seven image datasets as listed in Table~\ref{table:dataset}, which can be grouped into three categories. The first two datasets represent digits ($0$--$9$): MNIST~\cite{lecun1998gradient} for handwritten digits and SVHN~\cite{netzer2011reading} for street-view house numbers. The next two datasets contain images of English characters and digits (`a'-`z', `A'-`Z', and `0'-`9'): FEMNIST~\cite{caldas2018leaf} for handwritten ones, and Chars74K~\cite{de2009character} for a mix of characters obtained from outdoor images, hand-written characters, and computer fonts. The last three datasets represent images of various objects: FASHION~\cite{xiao2017fashion} for fashion items, and CIFAR-10 and CIFAR-100~\cite{krizhevsky2009learning} for different types of objects (e.g., vehicles, animals).

\subsubsection{Data Partition Among Clients}

We partition data into clients in a non-i.i.d. manner as in realistic federated learning scenarios. 
For FEMNIST which is already partitioned by the writers, we consider the images from each writer as belonging to a separate client. 
For all other datasets, the data are distributed into  clients so that each client has only one class (label) of data from that dataset. The clients and labels are associated randomly, so that the number of clients for each label differ by at most one. For all clients with the same label, the data with this label are partitioned into clients randomly.
When multiple datasets are mixed together (the open-set noise setting), different clients have different proportions of samples from each dataset, resulting in different amount of data samples in total.
We have $N=370$ clients in any experiment involving FEMNIST which is obtained using the default value in FEMNIST for non-i.i.d.  partition\footnote{\url{https://github.com/TalwalkarLab/leaf/tree/master/data/femnist}}.
For experiments without FEMNIST, we assume $N=100$ clients.

\begin{figure*}
    \centering
    
     \begin{subfigure}[b]{0.7\linewidth}
        \centering
        \includegraphics[width=1\linewidth]{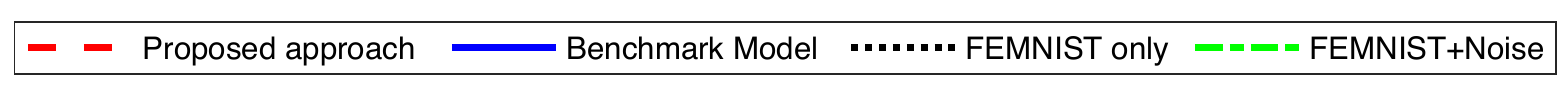}
    \end{subfigure}%
    
    \begin{subfigure}[b]{0.32\linewidth}
        \centering
        \includegraphics[width=1\linewidth]{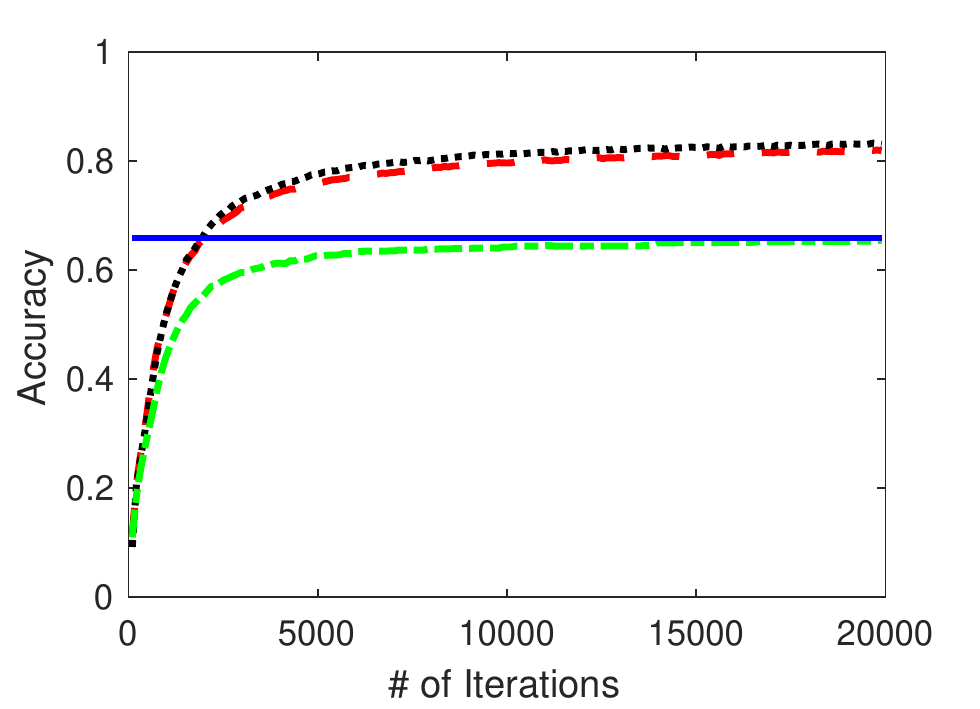}
        \caption{Noise: CHAR-74}
        \label{fig:varying-noise2}
    \end{subfigure}%
    \centering
    \begin{subfigure}[b]{0.32\linewidth}
        \centering
        \includegraphics[width=1\linewidth]{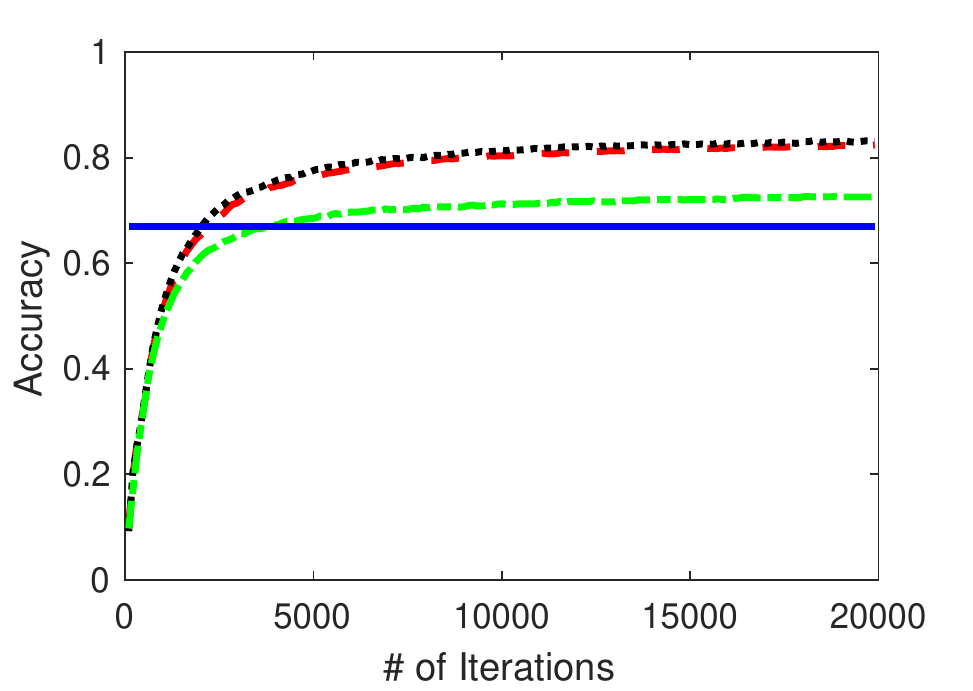}
        \caption{Noise: CIFAR-100}
        \label{fig:varying-noise4}
    \end{subfigure}%
    \begin{subfigure}[b]{0.32\linewidth}
        \centering
        \includegraphics[width=1\linewidth]{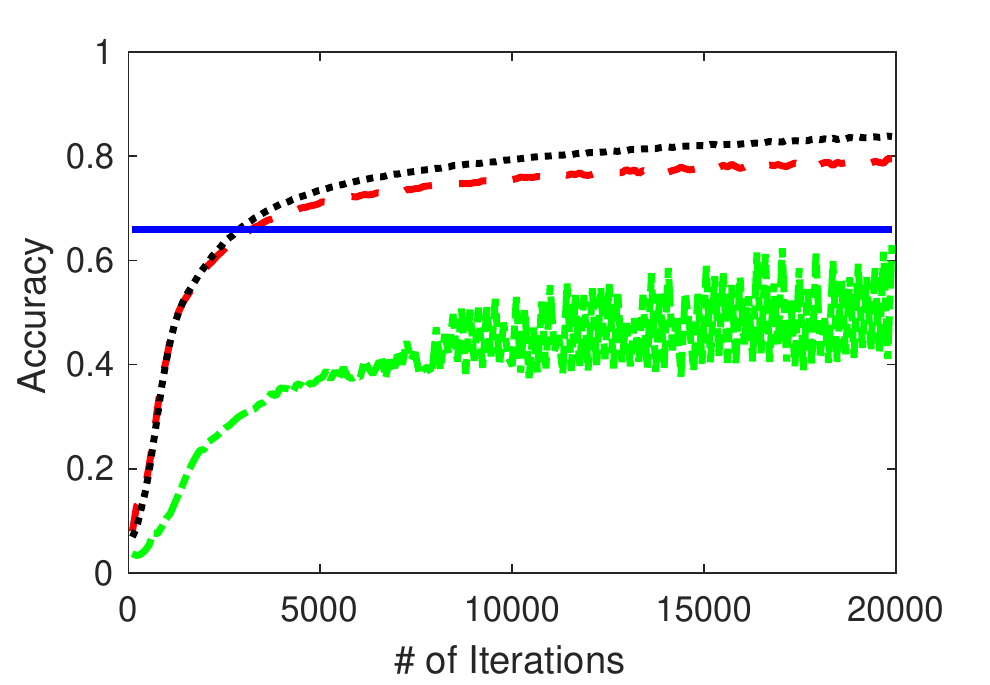}
        \caption{Noise: Closed-set}
        \label{fig:varying-noise6}
    \end{subfigure}%
    \centering

\caption{Classifying FEMNIST under different types of noise, when the amount of benchmark data is $3 \%$ of the original data.}
\label{fig:varying-noise}
\end{figure*}

\begin{figure*}
    \centering

    \begin{subfigure}[b]{0.7\linewidth}
        \centering
        \includegraphics[width=1\linewidth]{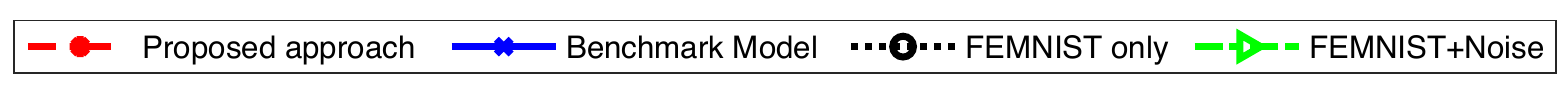}
    \end{subfigure}%
    
    \begin{subfigure}[b]{0.32\linewidth}
        \centering
        \includegraphics[width=1\linewidth]{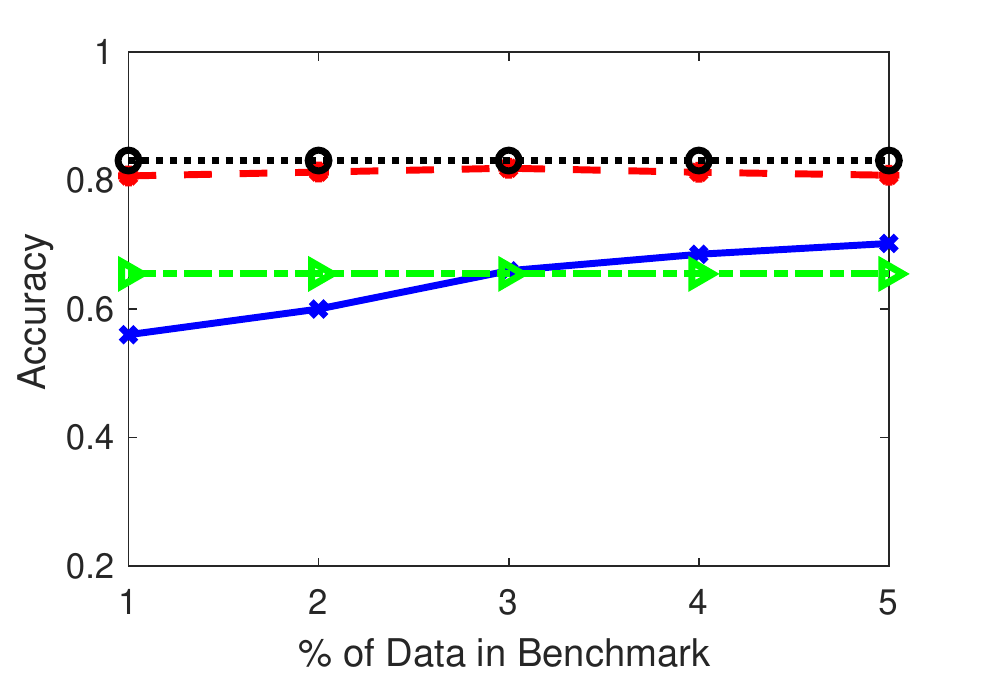}
        \caption{Noise: CHAR-74, Target: FEMNIST}
        \label{fig:varying-clean1}
    \end{subfigure}%
    \begin{subfigure}[b]{0.32\linewidth}
        \centering
        \includegraphics[width=1\linewidth]{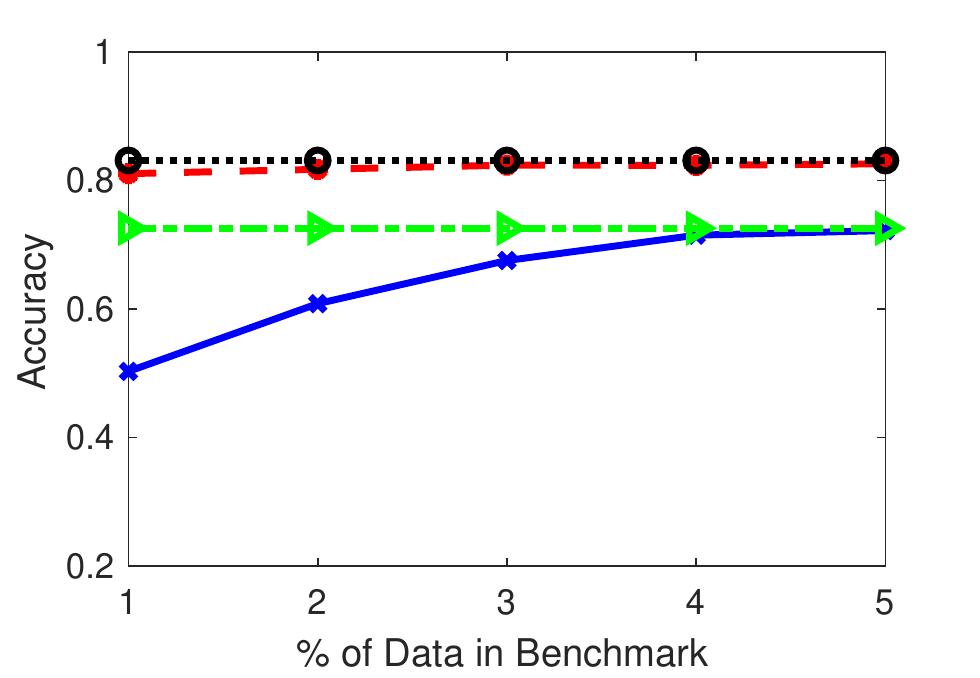}
        \caption{Noise: CIFAR-100, Target: FEMNIST}
        \label{fig:varying-clean2}
    \end{subfigure}%
    \begin{subfigure}[b]{0.32\linewidth}
        \centering
        \includegraphics[width=1\linewidth]{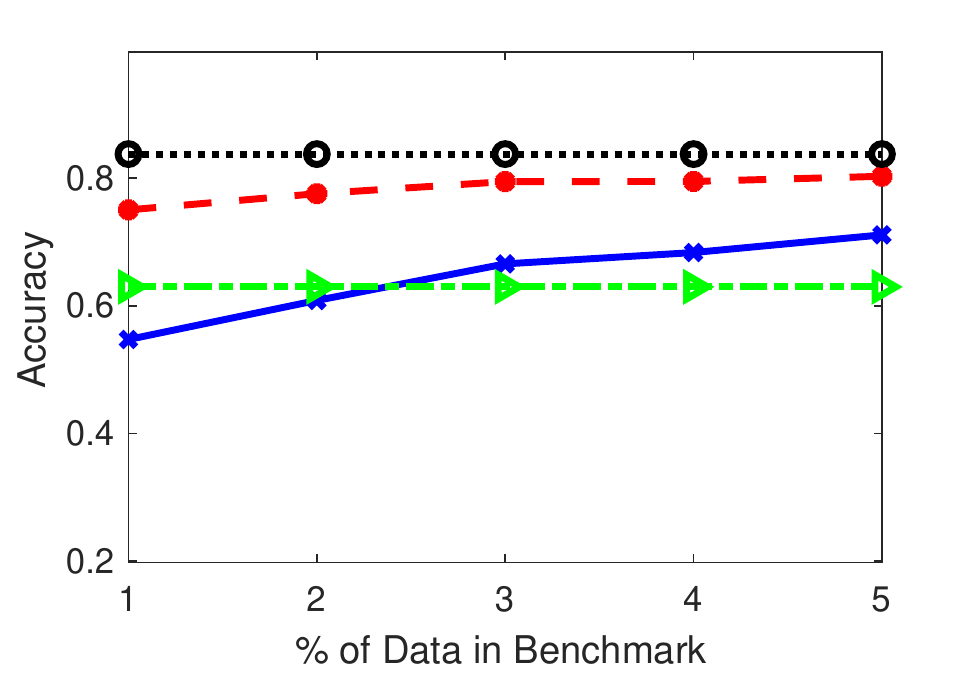}
        \caption{Noise: Closed-set, Target: FEMNIST}
        \label{fig:varying-clean3}
    \end{subfigure}%
    \centering
     
\caption{Varying size of benchmark data when classifying FEMNIST under different types of noise.}
\label{fig:varying-clean}
\end{figure*}

\subsubsection{Noisy Data}
\label{sec:noise}

Open-set noisy data are constructed by adding data from other datasets as noise to a given ``target'' dataset (i.e., the clean dataset for the model being trained in the task). We mix the noisy data samples to the target dataset, and preserve their labels in the mixed dataset\footnote{This is a common practice for simulating open-set noise \cite{wang2018iterative,hendrycks2018using}.}.
For example, we sample some data from SVHN (acting as noise, with labels 0--9) and add them to MNIST (acting as  the target dataset, with labels 0--9) while keeping the same label, e.g., data with label 0 in SVHN is mixed only with data with label 0 in MNIST. Even if the number of labels in the noise dataset is different from that of the target dataset, we apply the noise only to the labels that are common to both datasets; e.g., CIFAR-10's data with 10 labels are added to the corresponding first 10 classes of FEMNIST, or CIFAR-100's first 62 classes of data are mixed to the corresponding classes of FEMNIST. 
Additionally, in order to mix different datasets and train them together, we transform the open-set noise data such that their dimensions are the same as that of the target dataset (e.g., when training a classifier for SVHN with MNIST as noise, we transform MNIST to color images, and resize them to $32 \times 32$ pixels). 

In closed-set noise settings where the noise and the clean data both belong to a single dataset, which can be seen as a special case of the open-set noise setting, a subset of the dataset is mislabeled from one class to another.

\subsubsection{Benchmark Data} \label{subsec:benchmarkData}
The benchmark dataset used to build the benchmark model is obtained by sampling a certain percentage of data from the original (clean) datasets. For all datasets other than FEMNIST, the benchmark dataset is sampled randomly from the training data. For FEMNIST, which is pre-partitioned, the benchmark dataset only includes data from a small subset of partitions (clients) corresponding to individual writers.

\subsubsection{Model and Baseline Methods}

We use a convolutional neural network (CNN) as the classifier for all experiments. The CNN architecture is the same as the one used by \cite{wang2019adaptive}.
After data selection, we perform federated learning where each client $n$ only uses its selected data $\mathcal{F}_n$ (see Section~\ref{sec:data-filter}), using SGD with learning rate $\eta = 0.01$, number of local updates $\tau=10$, and a mini-batch size of $8\%$ of $\mathcal{F}_n$ for each client $n$ (see Section~\ref{section:federated}).

To evaluate our data selection method, we consider three baseline methods for comparison:
\begin{enumerate}
    \item Model trained only on the small benchmark dataset (referred to as the \textit{benchmark model});
    \item Model trained with federated learning using only the target dataset without noise at each client (i.e., the ideal case);
    \item Model trained with federated learning using all the (noisy) data at clients without data selection.
\end{enumerate}
For these baseline methods, we use the same training parameters as above.

\subsection{Results}

\subsubsection{Data Selection Performance (Different Types of Noise)}
\label{sec:mild-noise}

We first conduct experiments with different types of noise: (i) open-set noise from datasets in the same category, (ii) open-set noise from a different category, and (iii) closed-set noise in the same dataset. 
In particular, we use FEMNIST as the target dataset, and mix Chars74k into it as the same-category open-set noise, and CIFAR-100 as the different-category open-set noise\footnote{We observe similar results from other combinations of target and noise datasets, which are omitted in this paper for brevity.}. In both cases, we produce the data in a $1$:$1$ target to noise ratio in terms of the number of samples. For the closed-set noise, we mislabel samples that have labels from $0$ to $31$ by adding +2 to the true label, in $75\%$ of the clients, resulting in approximately $37.5\%$ of the total dataset being mislabeled. The benchmark dataset is generated by sampling up to $5\%$ of FEMNIST (see Section~\ref{subsec:benchmarkData}). 

\begin{figure}
    \centering
  \begin{subfigure}[b]{0.6\columnwidth}
        \centering
        \includegraphics[width=0.8\linewidth]{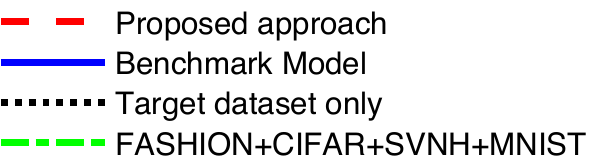}
    \end{subfigure}
    \begin{subfigure}[b]{1\columnwidth}
        \centering
        \includegraphics[width=0.7\linewidth]{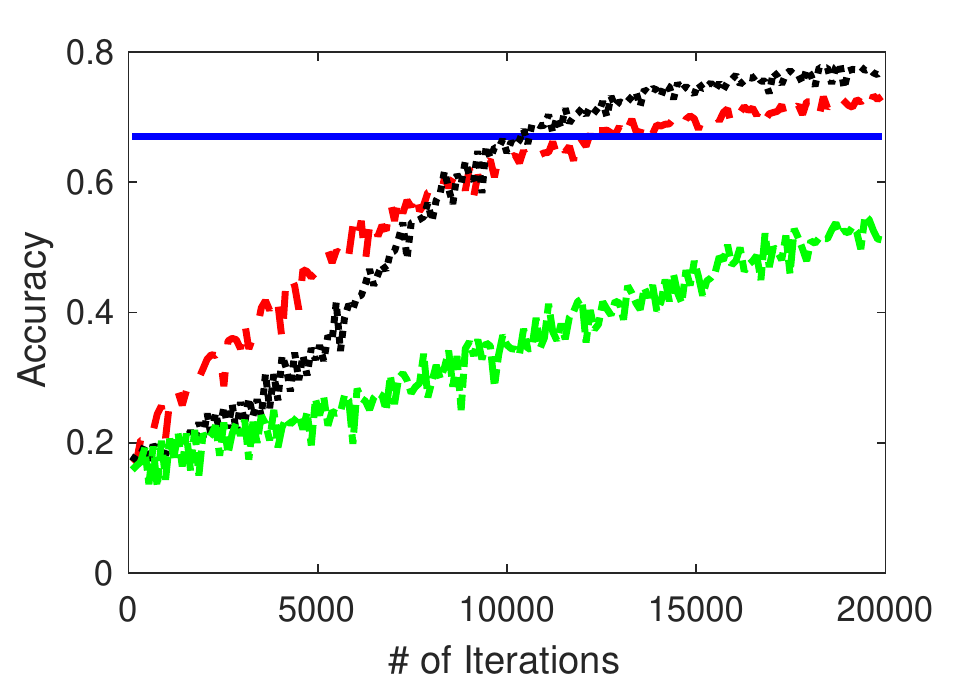}
        \caption{Target:  SVHN}
        \label{fig:strong-noise2}
        \hspace{1cm}
    \end{subfigure}%
    
    \begin{subfigure}[b]{1\columnwidth}
        \centering
        \includegraphics[width=0.7\linewidth]{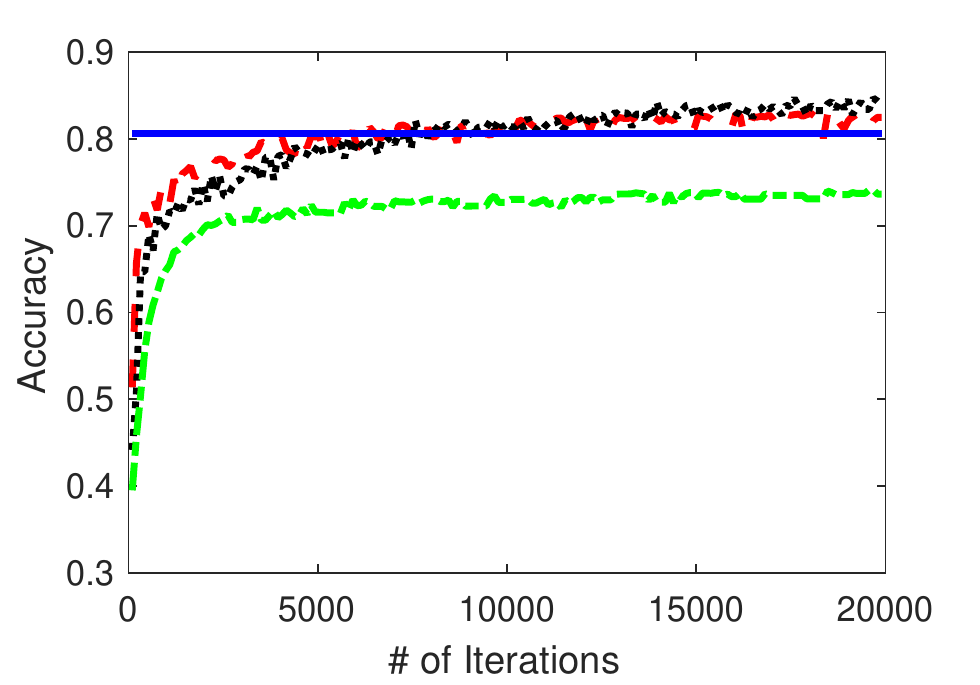}
        \caption{Target:  FASHION}
        \label{fig:strong-noise4}
    \end{subfigure}
\caption{Strong open-set noise scenario, when the amount of benchmark data is $3\%$ of the original data.}
\label{fig:strong-noise}
\end{figure}

Figure~\ref{fig:varying-noise} shows the accuracy achieved by the four models (obtained by our method and the three baselines) on the test data when  they are trained with the above three noisy data settings.
The performance of the benchmark model is shown as a constant in Figure~\ref{fig:varying-noise} (and also in Figure~\ref{fig:strong-noise} later), because it is trained at the MR before federated learning starts.
We see that, in all cases, our approach always performs very close to the ideal case baseline (``FEMINIST only''), and significantly better than the benchmark model and the one trained with the entire dataset with both target and noise. This shows the robustness of our approach to both open-set and closed-set noise.

\begin{figure}
    \centering
     \begin{subfigure}[b]{1\columnwidth}
        \centering
        \includegraphics[width=0.5\linewidth]{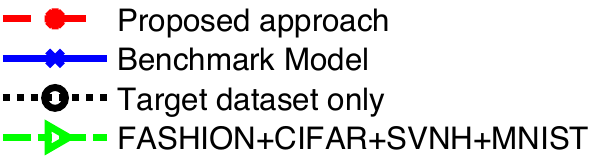}
    \end{subfigure}%
    
     \begin{subfigure}[b]{1\columnwidth}
        \centering
        \includegraphics[width=0.7\linewidth]{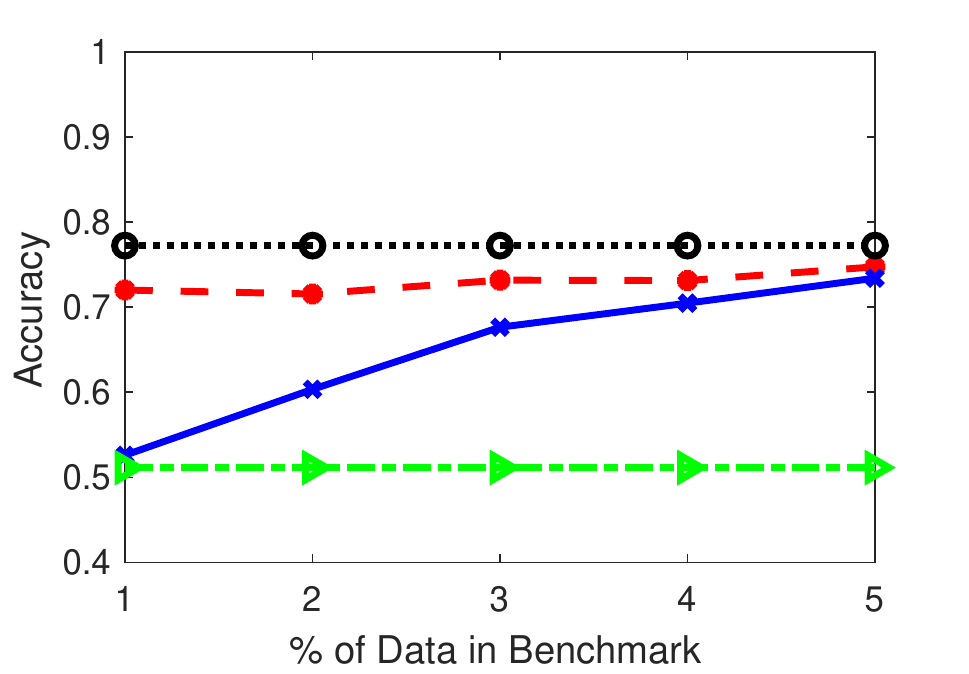}
        \caption{Target: SVHN}
        \label{fig:varying-clean-strong1}
    \end{subfigure}%
    
    \begin{subfigure}[b]{1\columnwidth}
        \centering
        \includegraphics[width=0.7\linewidth]{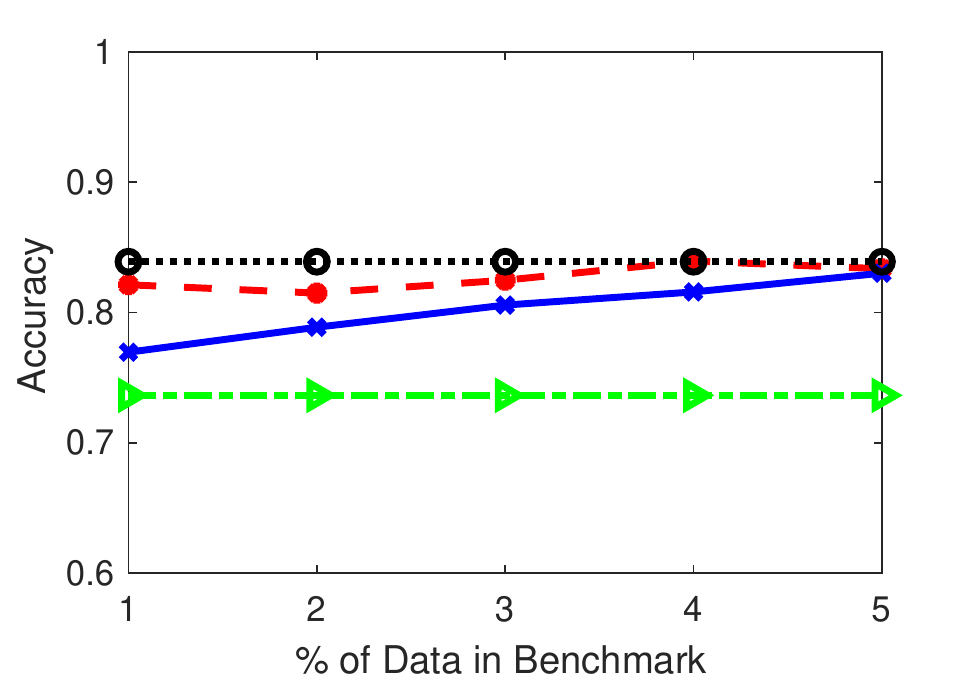}
        \caption{Target:  FASHION}
        \label{fig:varying-clean-strong2}
    \end{subfigure}%
\caption{Varying size of benchmark data with strong noise.} 
\label{fig:varying-clean-strong}
\end{figure}

Figure~\ref{fig:varying-clean} shows the results of repeating the same experiments but with the amount of benchmark data varied from $1\%$ to $5\%$ of the original dataset. The results, shown as the accuracy on the test data when federated learning has converged, clearly indicate that the model built with our data selection method performs very close to the ideal-case baseline, while outperforming the other two baselines (benchmark model and noisy data model), for all sizes of the benchmark dataset. We also see that the performance of the benchmark model increases with the (clean) benchmark dataset size as expected, while the performance of our approach remains nearly constant. This shows that our approach works well even with a very small amount of benchmark data (such as $1\%$ of the original data).

\subsubsection{Data Selection Performance (Strong Noise)}
\label{sec:strong-open-noise}

We then conduct experiments to assess our data selection method when we further increase the level of the noise such that $75\%$ of the training data are noise.
Figure~\ref{fig:strong-noise} shows the results (testing accuracy) for the cases when (i) SVHN is the target dataset with CIFAR-10, MNIST, and FASHION as noise (Figure~\ref{fig:strong-noise2}), and (ii) FASHION is the target dataset with CIFAR-10, SVHN, and MNIST as noise (Figure~\ref{fig:strong-noise4}). 
Additionally, Figure~\ref{fig:varying-clean-strong} shows results for varying sizes of benchmark dataset.

The overall trend in the performance is similar to what is observed with mild noise levels in Figures~\ref{fig:varying-noise} and \ref{fig:varying-clean}. Our approach achieves a model accuracy very close to the ideal-case baseline, while significantly outperforming the case without data selection. The benchmark model in this case performs reasonably well, too, since a relatively small amount of training data for these two target datasets (SVHN and FASHION) is generally sufficient to train a model achieving good accuracy. However, as observed in Figure~\ref{fig:varying-clean-strong}, the benchmark models still suffer when the benchmark dataset is very small, whereas our data selection method performs well, indicating that our method enables effective federated learning even under strong noise levels and small benchmark dataset. 

\subsubsection{Completion Time}
\begin{figure}
    \centering

    \begin{subfigure}[b]{1\columnwidth}
        \centering
        \includegraphics[width=0.7\linewidth]{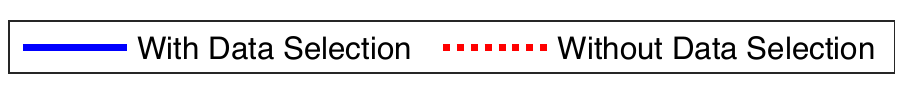}
    \end{subfigure}%
    
     \begin{subfigure}[b]{0.5\columnwidth}
        \centering
        \includegraphics[width=1\linewidth]{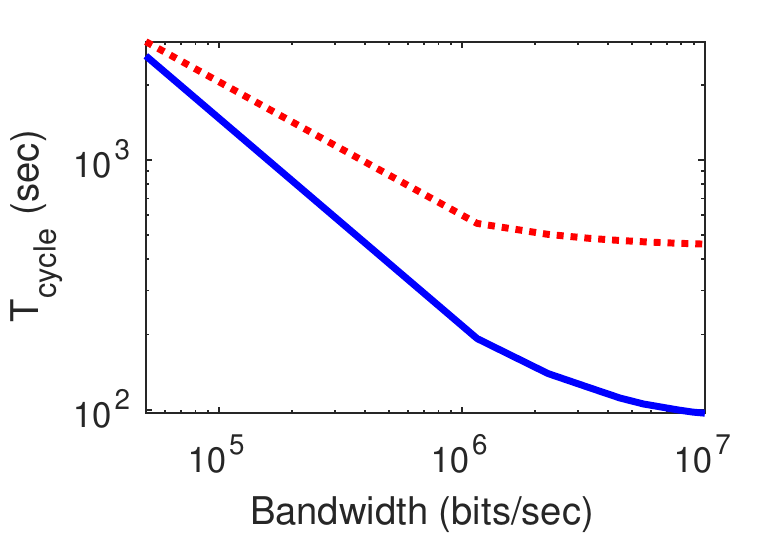}
    \end{subfigure}

\caption{Comparison of the average time of a cycle with and without data selection.} 
\label{fig:timing-comparison}
\vspace{-0.05in}
\end{figure}

We also compare the average completion time  of a federated learning ``cycle'' with and without our data selection approach, using SVNH as the target dataset as in Figures~\ref{fig:strong-noise2} and \ref{fig:varying-clean-strong1}. Here, a cycle of federated learning is defined as the time duration for all the clients $n=1,2,...,N$ to \emph{download} the model parameter vector $\theta_n^{(k)}(t)$, \emph{compute} $\tau$ steps of model update according to (\ref{eq:local}) on local data, and \emph{upload} the new $\theta_n^{(k)}(t+\tau)$ to the server. 
We study the time of a cycle using real measurements of the computation time on Raspberry Pi 4 devices and simulated transmission times (for sending model parameters between clients and server) under different communication bandwidths.
The results are shown in Figure~\ref{fig:timing-comparison}. We see that data selection also reduces the training time, because by having only a subset of relevant data involved in each task, the mini-batch sizes at clients are smaller than involving all the data at clients, as we use a variable mini-batch size equal to a fixed percentage of the data size (see the discussion in Section~\ref{section:federated}). Note that even if we fix the mini-batch size, the overall time for each client to make a pass over all its data (i.e., one epoch) is still shorter when using data selection. 
\vspace{-0.05in}

\section{Conclusion}
\label{sec:conclusion}

In this paper, we have considered a challenge in federated learning where each client may have various types of local data with noisy labels. To overcome this challenge, we have proposed a method for selecting the subset of relevant data to be involved in a federated learning task. Through extensive experimental analysis using multiple real-world image datasets, we have demonstrated the effectiveness of our data selection method in strong open-set noise setting, and its advantages over multiple baseline approaches.

\bibliographystyle{IEEEtran}
\bibliography{ref}

\begin{thebibliography}{10}
\providecommand{\url}[1]{#1}
\csname url@samestyle\endcsname
\providecommand{\newblock}{\relax}
\providecommand{\bibinfo}[2]{#2}
\providecommand{\BIBentrySTDinterwordspacing}{\spaceskip=0pt\relax}
\providecommand{\BIBentryALTinterwordstretchfactor}{4}
\providecommand{\BIBentryALTinterwordspacing}{\spaceskip=\fontdimen2\font plus
\BIBentryALTinterwordstretchfactor\fontdimen3\font minus
  \fontdimen4\font\relax}
\providecommand{\BIBforeignlanguage}[2]{{%
\expandafter\ifx\csname l@#1\endcsname\relax
\typeout{** WARNING: IEEEtran.bst: No hyphenation pattern has been}%
\typeout{** loaded for the language `#1'. Using the pattern for}%
\typeout{** the default language instead.}%
\else
\language=\csname l@#1\endcsname
\fi
#2}}
\providecommand{\BIBdecl}{\relax}
\BIBdecl

\bibitem{mcmahan2016communication}
H.~B. McMahan, E.~Moore, D.~Ramage, S.~Hampson, and B.~A. y~Arcas,
  ``Communication-efficient learning of deep networks from decentralized
  data,'' in \emph{AISTATS}, 2016.

\bibitem{park2019wireless}
J.~Park, S.~Samarakoon, M.~Bennis, and M.~Debbah, ``Wireless network
  intelligence at the edge,'' \emph{Proceedings of the IEEE}, vol. 107, no.~11,
  pp. 2204--2239, 2019.

\bibitem{yang2019federated}
Q.~Yang, Y.~Liu, T.~Chen, and Y.~Tong, ``Federated machine learning: Concept
  and applications,'' \emph{ACM Transactions on Intelligent Systems and
  Technology (TIST)}, vol.~10, no.~2, p.~12, 2019.

\bibitem{li2019federated}
T.~Li, A.~K. Sahu, A.~Talwalkar, and V.~Smith, ``Federated learning:
  Challenges, methods, and future directions,'' \emph{arXiv preprint
  arXiv:1908.07873}, 2019.

\bibitem{kairouz2019advances}
P.~Kairouz, H.~B. McMahan \emph{et~al.}, ``Advances and open problems in
  federated learning,'' \emph{arXiv preprint arXiv:1912.04977}, 2019.

\bibitem{blanchard2017machine}
P.~Blanchard, R.~Guerraoui, J.~Stainer \emph{et~al.}, ``Machine learning with
  adversaries: Byzantine tolerant gradient descent,'' in \emph{Advances in
  Neural Information Processing Systems}, 2017, pp. 119--129.

\bibitem{fung2018mitigating}
C.~Fung, C.~J. Yoon, and I.~Beschastnikh, ``Mitigating sybils in federated
  learning poisoning,'' \emph{arXiv preprint arXiv:1808.04866}, 2018.

\bibitem{wang2019beyond}
Z.~Wang, M.~Song, Z.~Zhang, Y.~Song, Q.~Wang, and H.~Qi, ``Beyond inferring
  class representatives: User-level privacy leakage from federated learning,''
  in \emph{INFOCOM}.\hskip 1em plus 0.5em minus 0.4em\relax IEEE, 2019, pp.
  2512--2520.

\bibitem{wang2018iterative}
Y.~Wang, W.~Liu, X.~Ma, J.~Bailey, H.~Zha, L.~Song, and S.-T. Xia, ``Iterative
  learning with open-set noisy labels,'' in \emph{CVPR}, 2018, pp. 8688--8696.

\bibitem{patrini2017making}
G.~Patrini, A.~Rozza, A.~Krishna~Menon, R.~Nock, and L.~Qu, ``Making deep
  neural networks robust to label noise: A loss correction approach,'' in
  \emph{CVPR}, 2017, pp. 1944--1952.

\bibitem{han2018co}
B.~Han, Q.~Yao, X.~Yu, G.~Niu, M.~Xu, W.~Hu, I.~Tsang, and M.~Sugiyama,
  ``Co-teaching: Robust training of deep neural networks with extremely noisy
  labels,'' in \emph{Advances in Neural Information Processing Systems}, 2018,
  pp. 8527--8537.

\bibitem{ghosh2017robust}
A.~Ghosh, H.~Kumar, and P.~Sastry, ``Robust loss functions under label noise
  for deep neural networks,'' in \emph{AAAI}, 2017.

\bibitem{veit2017learning}
A.~Veit, N.~Alldrin, G.~Chechik, I.~Krasin, A.~Gupta, and S.~Belongie,
  ``Learning from noisy large-scale datasets with minimal supervision,'' in
  \emph{CVPR}, 2017, pp. 839--847.

\bibitem{vahdat2017toward}
A.~Vahdat, ``Toward robustness against label noise in training deep
  discriminative neural networks,'' in \emph{Advances in Neural Information
  Processing Systems}, 2017, pp. 5596--5605.

\bibitem{hendrycks2018using}
D.~Hendrycks, M.~Mazeika, D.~Wilson, and K.~Gimpel, ``Using trusted data to
  train deep networks on labels corrupted by severe noise,'' in \emph{Advances
  in Neural Information Processing Systems}, 2018, pp. 10\,456--10\,465.

\bibitem{yu2019does}
X.~Yu, B.~Han, J.~Yao, G.~Niu, I.~Tsang, and M.~Sugiyama, ``How does
  disagreement help generalization against label corruption?'' in \emph{ICML},
  2019, pp. 7164--7173.

\bibitem{lee2019robust}
K.~Lee, S.~Yun, K.~Lee, H.~Lee, B.~Li, and J.~Shin, ``Robust inference via
  generative classifiers for handling noisy labels,'' \emph{arXiv preprint
  arXiv:1901.11300}, 2019.

\bibitem{Bergman2020Classification-Based}
\BIBentryALTinterwordspacing
L.~Bergman and Y.~Hoshen, ``Classification-based anomaly detection for general
  data,'' in \emph{International Conference on Learning Representations}, 2020.
  [Online]. Available: \url{https://openreview.net/forum?id=H1lK_lBtvS}
\BIBentrySTDinterwordspacing

\bibitem{wang2019adaptive}
S.~Wang, T.~Tuor, T.~Salonidis, K.~K. Leung, C.~Makaya, T.~He, and K.~Chan,
  ``Adaptive federated learning in resource constrained edge computing
  systems,'' \emph{IEEE Journal on Selected Areas in Communications}, vol.~37,
  no.~6, pp. 1205--1221, 2019.

\bibitem{tuor2018distributed}
T.~Tuor, S.~Wang, K.~K. Leung, and K.~Chan, ``Distributed machine learning in
  coalition environments: overview of techniques,'' in \emph{FUSION}.\hskip 1em
  plus 0.5em minus 0.4em\relax IEEE, 2018, pp. 814--821.

\bibitem{massey1951kolmogorov}
F.~J. Massey~Jr, ``The kolmogorov-smirnov test for goodness of fit,''
  \emph{Journal of the American statistical Association}, vol.~46, no. 253, pp.
  68--78, 1951.

\bibitem{lecun1998gradient}
Y.~LeCun, L.~Bottou, Y.~Bengio, and P.~Haffner, ``Gradient-based learning
  applied to document recognition,'' \emph{Proceedings of the IEEE}, vol.~86,
  no.~11, pp. 2278--2324, 1998.

\bibitem{netzer2011reading}
Y.~Netzer, T.~Wang, A.~Coates, A.~Bissacco, B.~Wu, and A.~Y. Ng, ``Reading
  digits in natural images with unsupervised feature learning,'' 2011.

\bibitem{caldas2018leaf}
S.~Caldas, P.~Wu, T.~Li, J.~Kone{\v{c}}n{\`y}, H.~B. McMahan, V.~Smith, and
  A.~Talwalkar, ``Leaf: A benchmark for federated settings,'' \emph{arXiv
  preprint arXiv:1812.01097}, 2018.

\bibitem{de2009character}
T.~E. De~Campos, B.~R. Babu, M.~Varma \emph{et~al.}, ``Character recognition in
  natural images.'' \emph{VISAPP (2)}, vol.~7, 2009.

\bibitem{xiao2017fashion}
H.~Xiao, K.~Rasul, and R.~Vollgraf, ``Fashion-mnist: a novel image dataset for
  benchmarking machine learning algorithms,'' \emph{arXiv preprint
  arXiv:1708.07747}, 2017.

\bibitem{krizhevsky2009learning}
A.~Krizhevsky, G.~Hinton \emph{et~al.}, ``Learning multiple layers of features
  from tiny images,'' Citeseer, Tech. Rep., 2009.

\end{thebibliography}

\end{document}